\documentclass[conference]{IEEEtran}
\IEEEoverridecommandlockouts
\usepackage{cite}
\usepackage{amsmath,amssymb,amsfonts}
\usepackage{algorithmic}
\usepackage{graphicx}
\usepackage{url}
\usepackage{hyperref}

\usepackage{lipsum} 
\usepackage{setspace} 
\usepackage{textcomp}
\usepackage{xcolor}
\def\BibTeX{{\rm B\kern-.05em{\sc i\kern-.025em b}\kern-.08em
    T\kern-.1667em\lower.7ex\hbox{E}\kern-.125emX}}
  
\begin{document}
\title{Enhancing Bidirectional Sign Language Communication:
Integrating YOLOv8 and NLP for Real-Time Gesture Recognition \& Translation}

\makeatletter
\newcommand{\linebreakand}{%
  \end{@IEEEauthorhalign}
  \hfill\mbox{}\par
  \mbox{}\hfill\begin{@IEEEauthorhalign}
}

\author{
  \IEEEauthorblockN{Hasnat Jamil Bhuiyan}
  \IEEEauthorblockA{\textit{Department of CSE} \\
    \textit{BRAC Univeristy}\\
    Dhaka, Bangladesh \\
    hasnat.jamil.bhuiyan@g.bracu.ac.bd}
  \and
  \IEEEauthorblockN{Mubtasim Fuad Mozumder}
  \IEEEauthorblockA{\textit{Department of CSE} \\
    \textit{BRAC Univeristy}\\
    Dhaka, Bangladesh \\
    mubtasim.fuad.mozumder@g.bracu.ac.bd}
  \and
  \IEEEauthorblockN{Md. Rabiul Islam Khan}
  \IEEEauthorblockA{\textit{Department of CSE} \\
    \textit{BRAC Univeristy}\\
    Dhaka, Bangladesh \\
    rabiul.islam.khan@g.bracu.ac.bd}
  \linebreakand 
  \IEEEauthorblockN{Md. Sabbir Ahmed}
  \IEEEauthorblockA{\textit{Department of CSE} \\
    \textit{BRAC Univeristy}\\
    Dhaka, Bangladesh \\
    sabbir.ahmed@bracu.ac.bd}
  \and
  \IEEEauthorblockN{Nabuat Zaman Nahim}
  \IEEEauthorblockA{\textit{Department of CSE} \\
    \textit{BRAC Univeristy}\\
    Dhaka, Bangladesh \\
    nabuat.zaman@bracu.ac.bd}
}

\maketitle

\begin{abstract}

The primary concern of this research is to take American Sign Language (ASL) data through real time camera footage and be able to convert the data and information into text. Adding to that,we are also putting focus on creating a framework that can also convert text into sign language in real time which can help us break the language barrier for the people who are in need. In this work, for recognising American Sign Language (ASL), we have used the You Only Look Once(YOLO) model and Convolutional Neural Network (CNN) model. YOLO model is run in real time and automatically extracts discriminative spatial-temporal characteristics from the raw video stream without the need for any prior knowledge, eliminating design flaws.The CNN model here is also run in real time for sign language detection. We  have introduced a novel method for converting text based input to sign language by making a framework that will take a sentence as input, identify keywords from that sentence and then show a video where sign language is performed with respect to the sentence given as input in real time.To the best of our knowledge, this is a rare study to demonstrate bidirectional sign language communication in real time in the American Sign Language (ASL).     
\end{abstract}


\section{\textbf{Introduction}}
 Sign language is a visual language expressed through physical movements instead of spoken words. Hands, eyes, facial emotions, and movement are all used as visual clues in this language.Like any other language, sign language has its own grammatical rules and linguistic structures. However, use of sign language still isn't a common practice across the globe. According to the World Health Organization (WHO), over 1.5 billion people in the world live with hearing loss \cite{b56}. Despite such a huge chunk of the global population suffering from the issue, only a minority of people know how to effectively communicate with them due to lack of awareness about sign language.There are more than 300 sign languages present in different regions of the world\cite{b56}. Since English is the language mostly used throughout the world, our work will be based on American Sign Language (ASL).
\\
The objectives of this work are breifly discussed below:
\begin{itemize}
    \item Creating a framework based on natural language processing (NLP) that can represent accurate sign language gestures and movements in real time when text input is provided, allowing hearing people to communicate with sign language users.
    \item Developing real time YOLO and CNN models that will recognize and translate sign language gestures into text. 
    \item Assessing the performance and accuracy of the suggested models by using real-world sign language data in rigorous testing and validation and taking into account aspects such as recognition accuracy, precision, translation quality, speed, and robustness.
    \item Evaluating the effectiveness of the suggested framework and models. 
\end{itemize}

\section{Related Works}

Here are some of the selected research works that inspired us to start working on our thesis topic in depth. First we will discuss the works related to text to sign language translation. In \cite{b2},a unique approach to translating English sentences to Indian Sign Language (ISL) is seen. Their proposed system takes a text input and converts it to ISL with the help of Lexical Functional Grammar (LFG). In \cite{b13}, an approach to transform Malayalam text to Indian Sign Language using animation for displaying is seen. Their system uses the Hamburg Notation System shortly known as HamNoSys for representing signs.Moreover, The authors in \cite{b14} used an approach for converting Greek text to Greek sign language. Translation is done using Vsigns, a web tool used for the synthesis of virtual signs. A system is proposed in \cite{b15} where text in English language is taken as input and then translated to HamNoSys representation. This is afterward converted into SiGML. A mapping system is used to link the text to the HamNoSys notation. This work may not be a direct example of text-to-sign language conversion which we expect. However, this provides us with insights into converting text to a signed notation system. Similar research works were done in \cite{b16} and \cite{b39}. Furthermore,in \cite{b41}, the authors proposed a machine translation model that takes both example based and rule-based Interlingua approaches to convert Arabic Text to Arabic Sign Language. Another work of Arabic Sign language for the deaf is presented in \cite{b44}. Adding to that,in \cite{b35}, a text-to-sign language conversion system for Indian Sign Language (ISL) is made which takes into account the language's distinctive alphabet and syntax. The system accepts input in alphabets or numerals only.
\\
Now, we will discuss the works related to sign language recognition. In \cite{b10}, the authors attempted to recognize the English alphabet and gestures in sign language and produced the accurate text version of the sign language using CNN and computer vision.In \cite{b9}, the researchers worked on reviewing multiple works on the recognition of Indian Sign Language (ISL). Their review of works on Histogram of Orientation Gradient(HOG), Histogram of Edge Frequency(HOEF) and Support Vector Machine (SVM) gave us meaningful insights. A similar work is seen in in \cite{b46}. Furthermore,in \cite{b11}, the authors worked on Indonesian sign language recognition was done using a YOLOv3 pre-trained model. They used both image and video data. The system’s performance was incredibly high during using image data and it was comparatively low while using video data. A similar work was done in \cite{b12} using  YOLOv3 model.From\cite{b17}, we learnt how the researchers worked on making an Italian sign language recognition system that identifies letters of the Italian alphabet in real-time using CNN and VGG-19. The work of the authors in \cite{b19} and in \cite{b36}, was insightful about how deep learning  works on sign language detection. Moreover in \cite{b42}, the authors developed an Android app that can convert real-time ASL input to speech or text where SVM was used to train the proposed model. Additionally in \cite{b28}, we were introduced to the idea of using surface electromyography (sEMG), accelerometer (ACC), and gyroscope (GYRO) sensors for subword recognition in Chinese Sign Language. Lastly in \cite{b38}, the authors worked on a sign language-to-voice turning system that uses image processing and machine learning.

\section{\textbf{Methodology}}
\noindent
Our work is mainly divided into two parts: 
\begin{enumerate}
    \item Text to sign language conversion
    \item Sign language to text conversion
\end{enumerate}
  
\subsection{\textbf{Text to Sign Language}}
We have implemented spoken language sentences to sign language translation using Natural Language Processing (NLP). The framework utilizes the Natural Language Toolkit (nltk) for part-of-speech tagging to identify the tense of the input sentence. A dictionary is used to hold verb counts for various tenses, and a list of predefined stop words is utilized for filtering. After the words have been lemmatized, particular tense-indicating words like "Before," "Will," or "Now" are added in accordance with the determined tense. 
\\
We have created our own dataset containing videos of sign language performed corresponding to words,alphabets and digits from 0 to 9. The most frequently used words in the English language are included in this dataset.  In total, there are 150 videos. The words will be added to a list after each word in the input sentence and its matching part-of-speech tag have been processed and lemmatized according to their POS tag. For every word in the list, its corresponding video file will be looked for in the dataset.The word will be split up into individual alphabets if there isn't a video available in the dataset. Finally, the processed words and the original sentence will be passed to a html template for rendering video demonstrating sign language performed. The primary benefit of this framework is that it does not require system modifications when a new word is added to the dataset.\\

We have developed a webpage to take input and translate the words  into sign language for this part. For this website, the Django framework was used, along with Javascript, HTML, and CSS for the front end. In this website, we have three main pages: HomePage, Sign up and Log-in. The homepage features a navigation bar with links to the Sign Up and Login pages. It also includes a video of sign language representation of the world “Hello”. Below the video, there is a call-to-action button labeled "Click to Start." This button directs the user to the Log-in page. The new users can create accounts in the sign up page using necessary credentials like username and password. Users are redirected to their personal dashboard, which includes a text input field, after logging in. It is possible to provide textual or audio input. The sentence is divided into key words when input is received. Users can submit audio input by speaking sentences aloud by using the microphone icon button located beneath the text input box. The website translates spoken words into text by using the JavaScript Web Speech API for speech recognition. Users can click the "Submit" button after they have typed their sentence or provided audio input. This action triggers the processing of the input, extracting keywords from the sentence, and generating the corresponding sign language representation. A list of keywords extracted from the submitted sentence is shown beneath the text input area. On the right side of the dashboard, there is a play/pause button. Users can click on that button to watch the videos demonstrating the sign language representation of each keyword. By seeing fig. \ref{fig:111}, fig. \ref{fig:1131} and  fig.\ref{fig:222} it can be perceived how the translation is being done. 

\noindent
To extract keywords from a sentence we are using Natural language processing (NLP). The ``Natural language toolkit (NLTK)" Python library is being used, which includes features like tokenization, stemming, POS-tagging, and more. We have employed word tokenization at first which splits a sentence by words.Tokenizing the sentence ``I am happy" will, for instance, result in an array of individual words like [``I", ``am", ``happy"].  Next, parts-of-speech (POS) tagging is being applied. For POS tagging, we are utilizing the NLTK library's pos tag function. After that, we have a list of stop words. In text processing, these are common terms that are frequently omitted in order to highlight the more significant words. Lastly, lemmatization is accomplished using WordNetLemmatizer function. Words are being lemmatized based on their pos tags.\\

\begin{figure}[!h]
    \centering
    \includegraphics[scale=0.39]{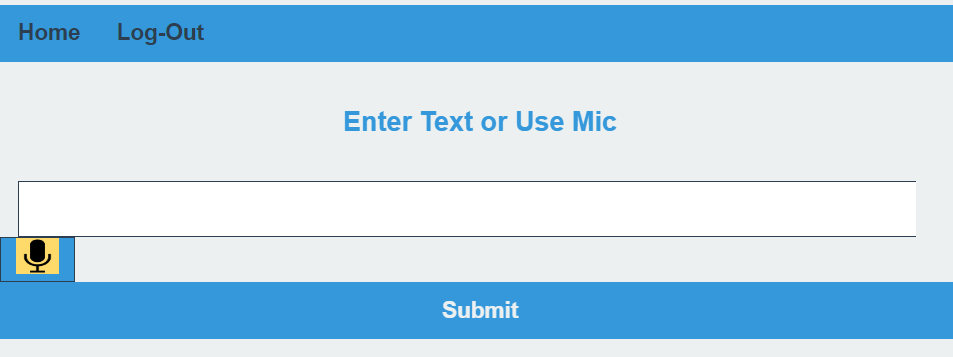}
    \caption{Taking input}
    \label{fig:111}
\end{figure}
\begin{figure}[!h]
    \centering
    \includegraphics[scale=0.35]{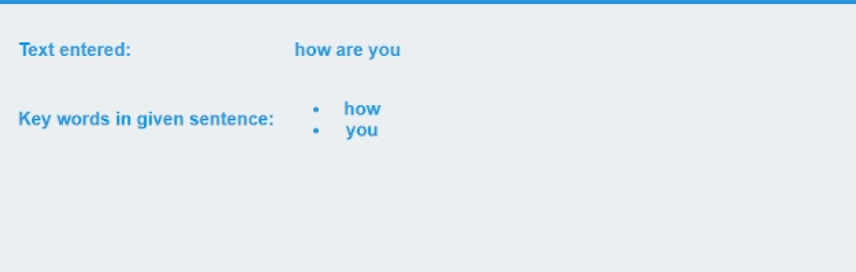}
    \caption{Key words in a sentence}
    \label{fig:1131}
\end{figure}

\begin{figure}[!h]
    \centering
    \includegraphics[scale=0.2]{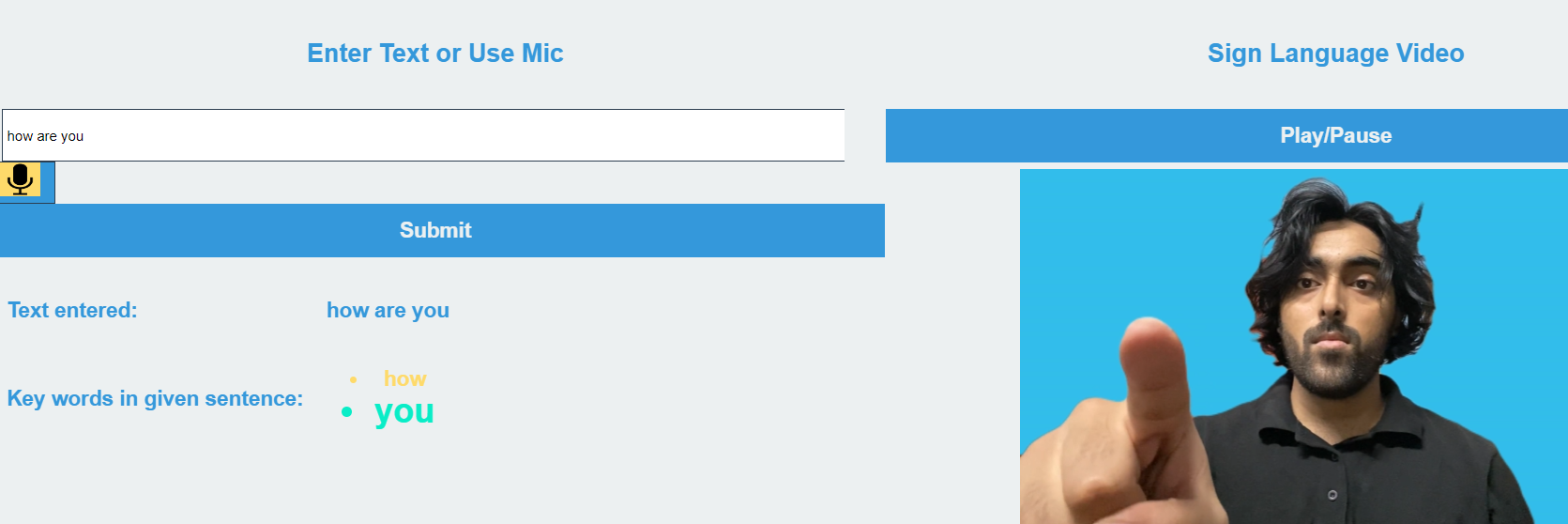}
    \caption{Snapshot of a sign language being performed}
    \label{fig:222}
\end{figure}
\subsection{\textbf{Sign Language to Text}}
\subsubsection{\textbf{YOLOv8 and v5 Models}}
For our sign language detection, we have developed YOLOv5 and v8 models, YOLOv8 is a state-of-the-art model in the ongoing time. This YOLO model is an upgraded version of its predecessors. It uses image and video data to train and can detect the gestures and other detection scenarios with exceptional accuracy. YOLOv8 uses a custom data loader mosaic which helps the data to be loaded into the model while training and testing the raw data. However, the data collection is more likely the same with v5 and v8 models as both have a similar architectural pattern. Both of these models are compatible with multiple data collection styles. We used bounding box prediction data style as it was more suitable for us as our hands move constantly and we will have to deal with more than one shape and aspect ratios. Therefore, we added our hand gestures with bounding boxes as it can be easily manipulated and the end result was satisfactory for both of the models.\\
Dataset preparation:\\
 We created our own database for ASL. Here we created classes for each sign. There are 22 classes in total. After creating images for each class and while resizing them at 640*480px, we annotated each of the data samples according to their specific class using the “labelImg” tool. 
\\The classes were split into 80[Train]:20[Test] ratio. The classes are: `Are you Hurt$\_$OR$\_$I am hurt$\_$', `Are You Thirsty$\_$OR$\_$I am Thirsty', `Be Careful',  `Did you eat$\_$OR$\_$I want to eat',  `Dont do that',  `Go There', `Hello',   `I am Cold$\_$OR$\_$Are you cold', `I am mad$\_$OR$\_$Why are you mad', `I am nervous', `I am Sad$\_$OR$\_$Why are you sad', `I am Sorry',  `I dont understand$\_$OR$\_$Do you understand', `I love you', `I Need Help',  `Need a bandage', `No', `Please$\_$OR$\_$Kindly', `Stop it$\_$Or$\_$Stop This', `Thank You',   `Why are you Crying,'`Yes'. In fig. \ref{fig:1171}, sample images of the classes are shown.

\begin{figure}[!h]
    \centering
    \includegraphics[scale=0.1]{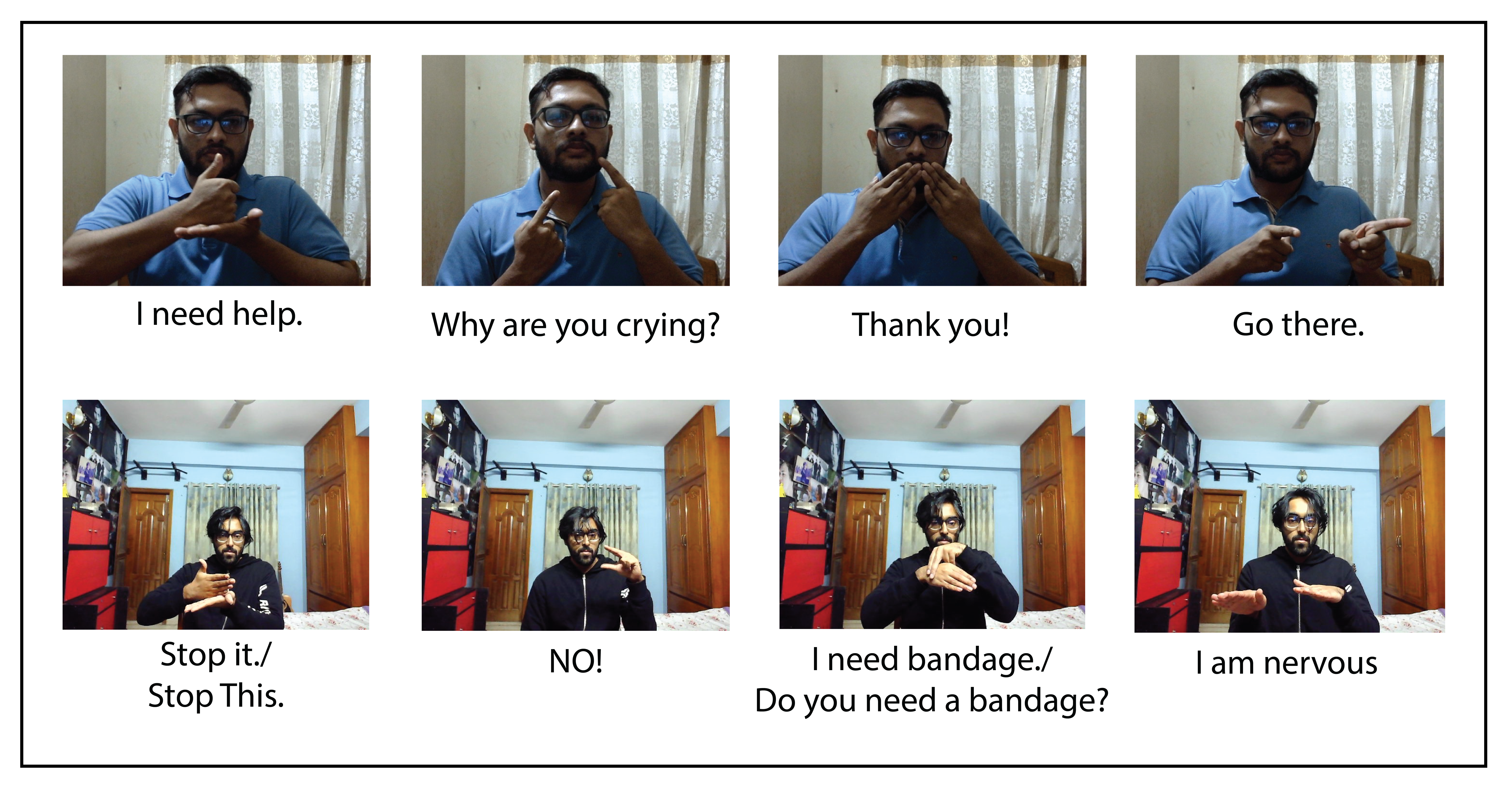}
    \caption{ Sample images of some of the classes.}
    \label{fig:1171}
\end{figure}
\noindent
Furthermore, most often the data of a single sentence have some movement including hand gestures. Therefore, we used samples which include movement for those specific sentences. The accuracy remained high as we used movement gestures. We took 80 image samples for each class in the train data and 20 image samples for each class in test data. Overall we got 2200 data samples, where we used 1760 samples for training and 440 samples for testing our data.

\noindent
\\
Implementation of YOLO v5 and v8 models:
\\

At first, we took the sample data one by one using a custom code after assigning the classes and numbers.Afterwards, we annotated each of the sample pictures with its corresponding class so that the actual annotation in real time may work. We preprocessed the same data which we collected and ran our decided epochs for both of the models. Our data was trained with 80 percent of the created dataset and it was validated with the other 20 percent. We took the best fit model we got from both of the training sessions and got the desired output.

\noindent
\\
Test results of YOLOv8 Model:\\

Here, 50 epochs was run for our YOLOv8 model with a total of 22 classes. In fig. \ref{fig:1122}, real time detection of some classes is shown and in fig. \ref{fig:112212} frames of real time moving sign language classes are shown.

\begin{figure}[!h]
    \centering
    \includegraphics[scale=0.42]{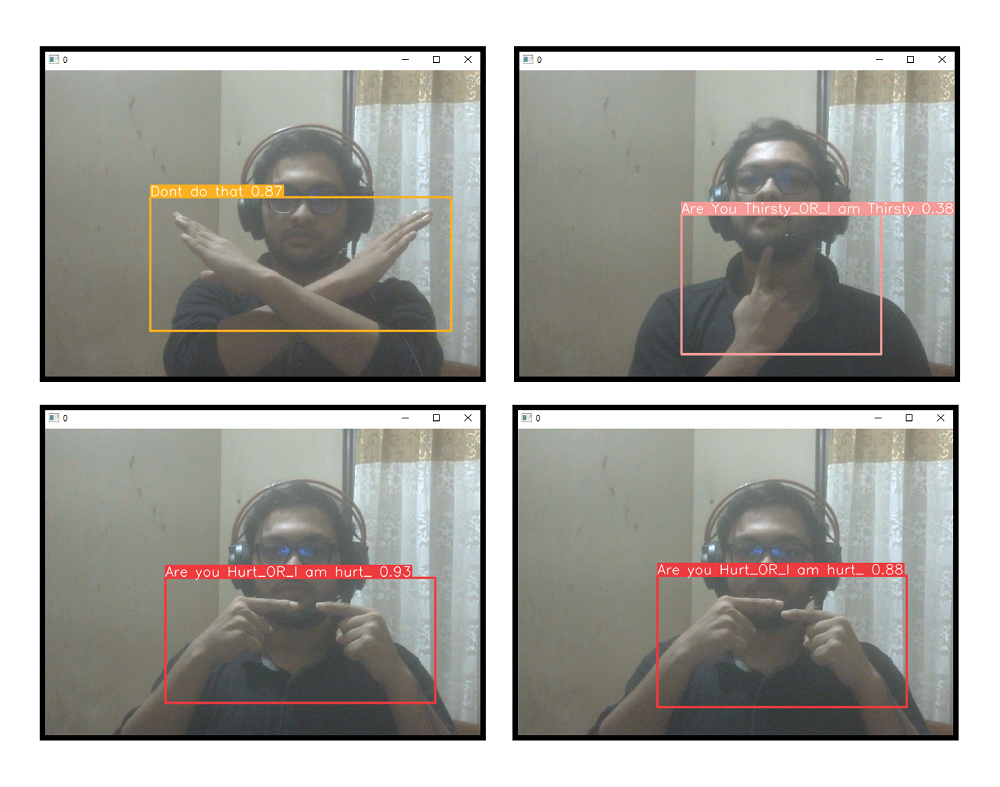}
    \caption{Real time detection output of our YOLOv8 model.}
    \label{fig:1122}
\end{figure}

\begin{figure}[!h]
    \centering
    \includegraphics[scale=.15]{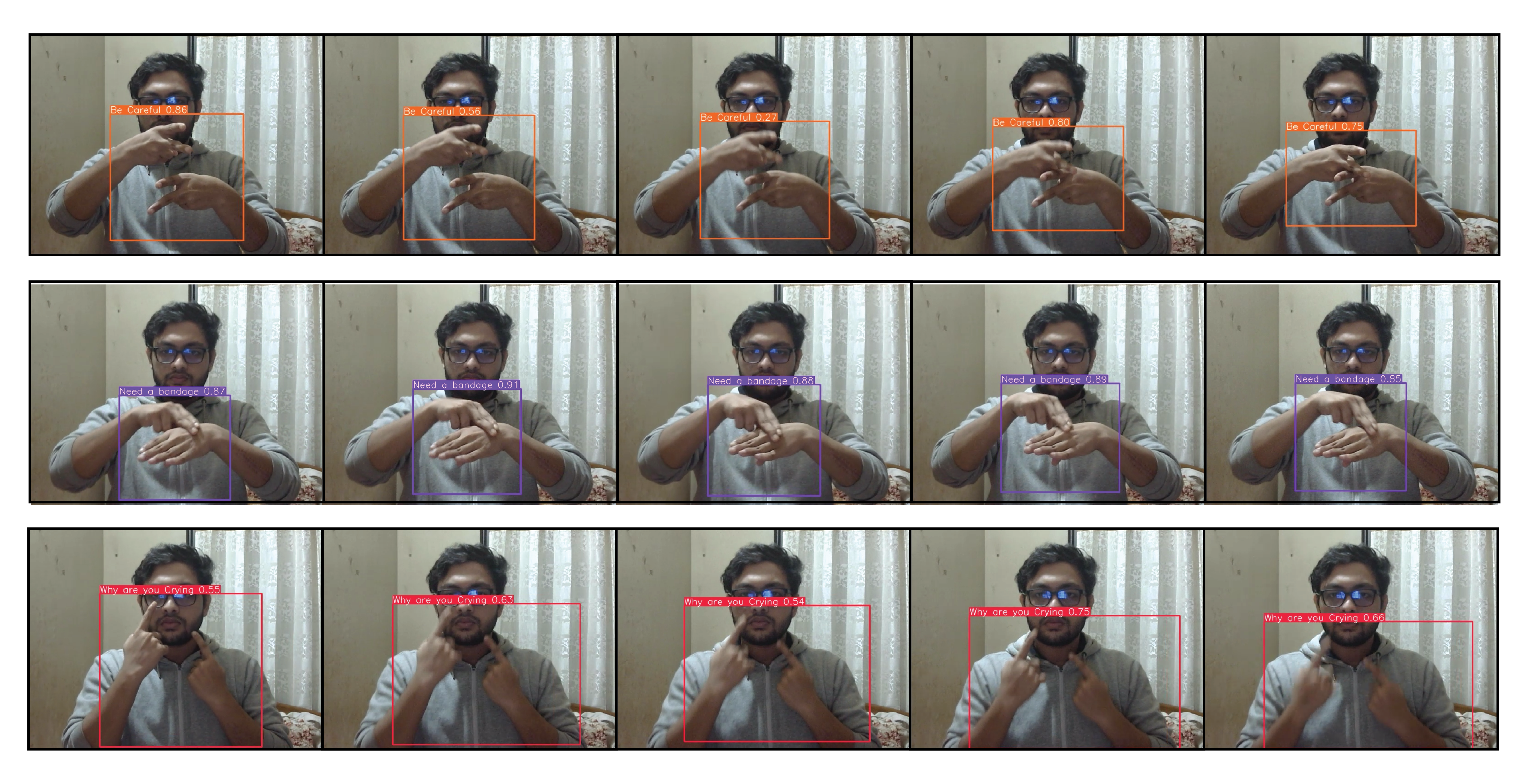}
    \caption{Frames of real time moving sign language classes from v8 model}
    \label{fig:112212}
\end{figure}

\begin{figure}[!h]
    \centering
    \includegraphics[scale=0.55]{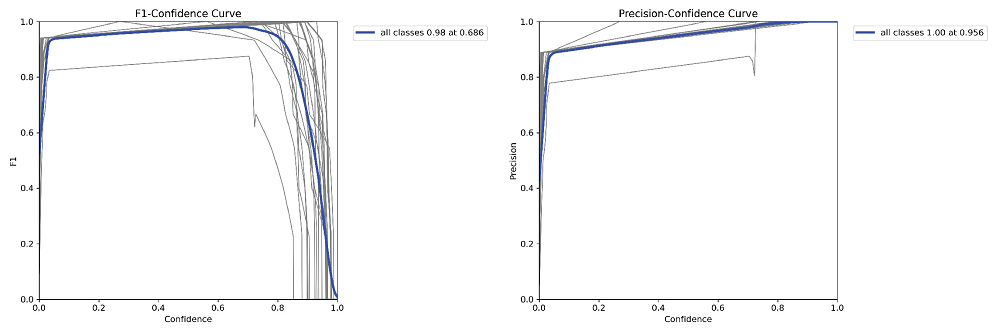}
    \caption{F1 confidence curve and precision graph of YOLOv8 model.}
    \label{fig:066}
\end{figure}
\noindent
From fig. \ref{fig:066}, it can be perceived that the confidence curve for the v8 model is 68.6 percent  and the precision level is 95.6 percent for all classes.

\begin{figure}[!h]
    \centering
    \includegraphics[scale=0.33]{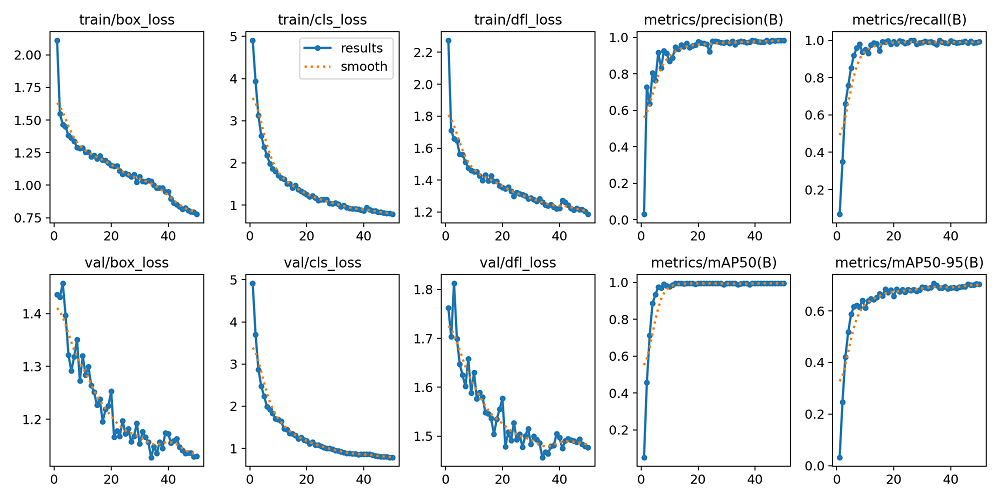}
    \caption{Validation graphs of YOLOv8 model.
}
    \label{fig:1142}
\end{figure}
\noindent
In fig.\ref{fig:1142} the training and validation box loss, class loss, object loss, precision metrics and recall metrics consecutively for the v8 model is represented. \\
\\
Test results of YOLOv5 Model:\\

\noindent
Here, we ran 100 epochs for our YOLOv5 model with the total of 22 classes.In fig. \ref{fig:1132}, real time detection of some classes is shown. 

\begin{figure}[!h]
    \centering
    \includegraphics[scale=0.335]{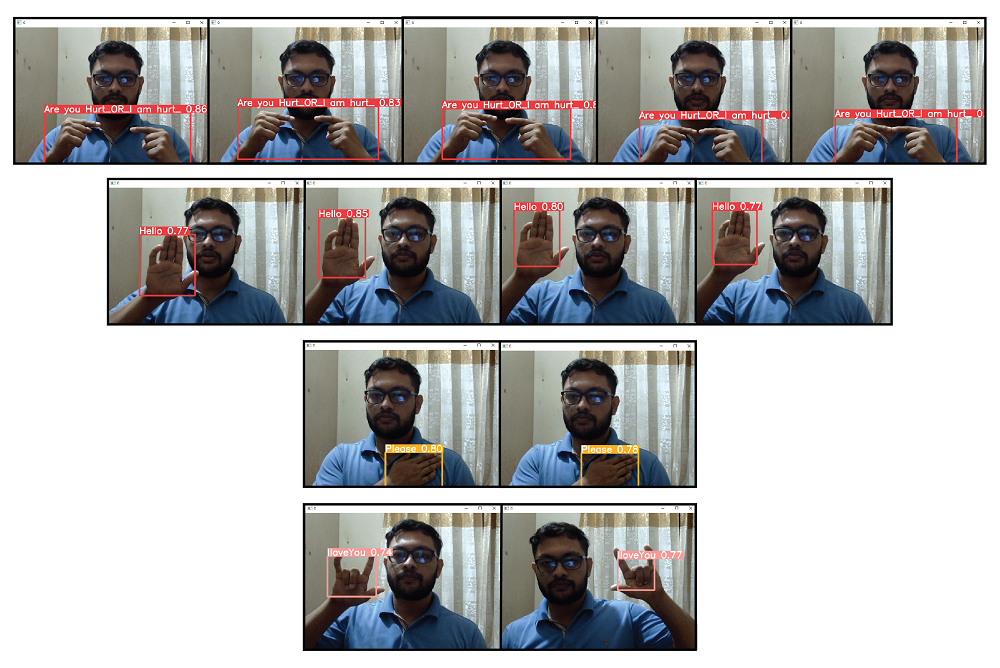}
    \caption{Real time detection output of our YOLOv5 model.
}
    \label{fig:1132}
\end{figure}

\begin{figure}[!h]
    \centering
    \includegraphics[scale=0.54]{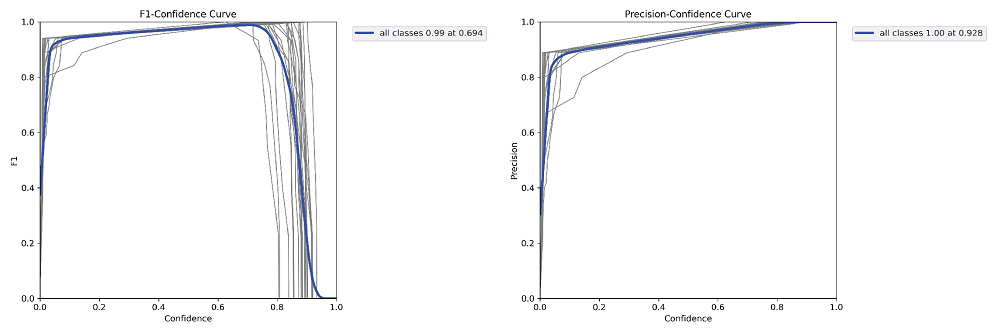}
    \caption{F1 confidence curve and precision graph of YOLOv5 model}
    \label{fig:18}
\end{figure}
\noindent
From fig. \ref{fig:18}, it is seen that the confidence curve for the v5 model is 69.4 percent and the precision level is 92.8 percent for all classes. 

\begin{figure}[!h]
    \centering
    \includegraphics[scale=0.35]{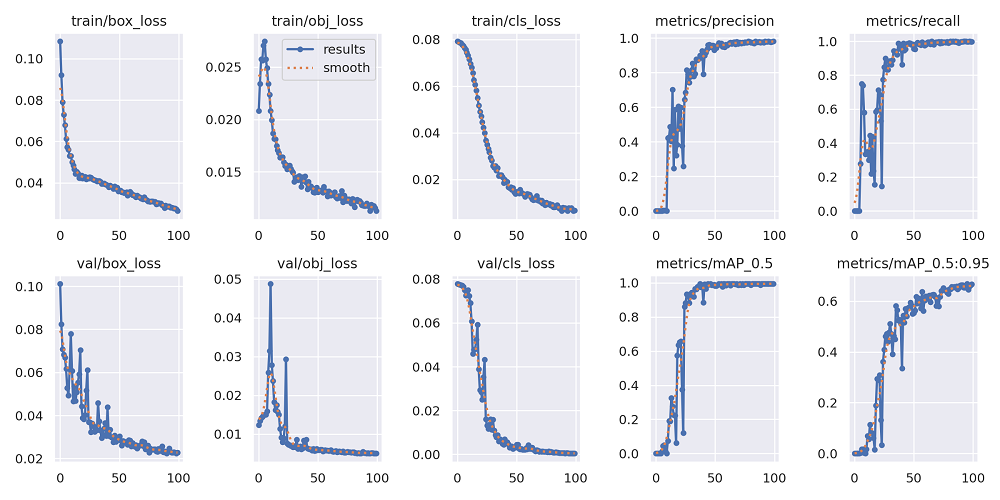}
    \caption{Validation graphs of YOLOv5 model.}
    \label{fig:1010}
\end{figure}

\noindent
From fig. \ref{fig:1010}, the training and validation box loss, object loss, class loss, precision metrics and recall metrics consecutively for the v5 model is shown. As we ran the model for 100 epochs. The curve is much smoother than the v8 model.

\subsubsection{\textbf{CNN Model}}
One kind of deep neural network that excels at image-based tasks is the convolutional neural network (CNN), which is especially useful for object detection.We have also done sign language detection with a simple CNN model implemented by python’s openCV library. \\

Dataset preparation:\\
Here, we created our own dataset as well.The dataset here is basically coordinate points taken from each frame. The coordinates here are taken for each and every class. Using the webcam and python’s opencv library, we collected data for each class. We have made hand sign gestures from different angles and at different parts of the frame while collecting data so that the model gets better trained and can detect the hand sign performed even if it has been performed from a different angle in the frame. The classes for the hand sign gestures where both hands are needed are provided with more data compared to the classes where only one hand is needed as identifying hand signs with both hands is difficult for this model. The model gets trained based on these coordinate points after preprocessing.The dataset has been made for the following 10 classes:'Hello there', 'I love you', 'I am sorry', 'Please', 'I need help', 'Go there', 'Why are you crying?', 'Be careful', ' Stop it', 'Don't do that'.      
\\
  
Implementation of CNN Model:\\
Now the working procedure of the model will be explained in detail. We used python’s opencv library to capture frames from webcam or any other camera. The frames are 2D matrices basically which hold pixel values and we have to identify in which part of the image the hands are being used to perform sign language. Using the mediapipe framework, we identified the joints in the user's hand which look like a skeleton-like structure, called landmarks.It was used to train a feed forward neural network which is simple, lightweight  and well enough to detect the user’s hand sign. The landmark is significant to our work. Because if we trained directly using the whole image then we would need a huge amount of training data in different environments and consider other elements like different hand and finger sizes.So, the skeleton-like structure simplifies our work a lot. Before training, the dataset is preprocessed. The objective in this step is to standardize and normalize the input data so that it is suitable to be fed to the CNN model. The dataset has been trained on the CNN model. The model was trained on 10 classes. The train and test ratio here was 75[train]:25[validation]. The model underwent 500 epochs with a batch size of 128 during the training process. The accuracy rate was 94.94 percent.

\noindent
Test results of CNN model:
\begin{figure}[!h]
    \centering
    \includegraphics[scale=0.06]{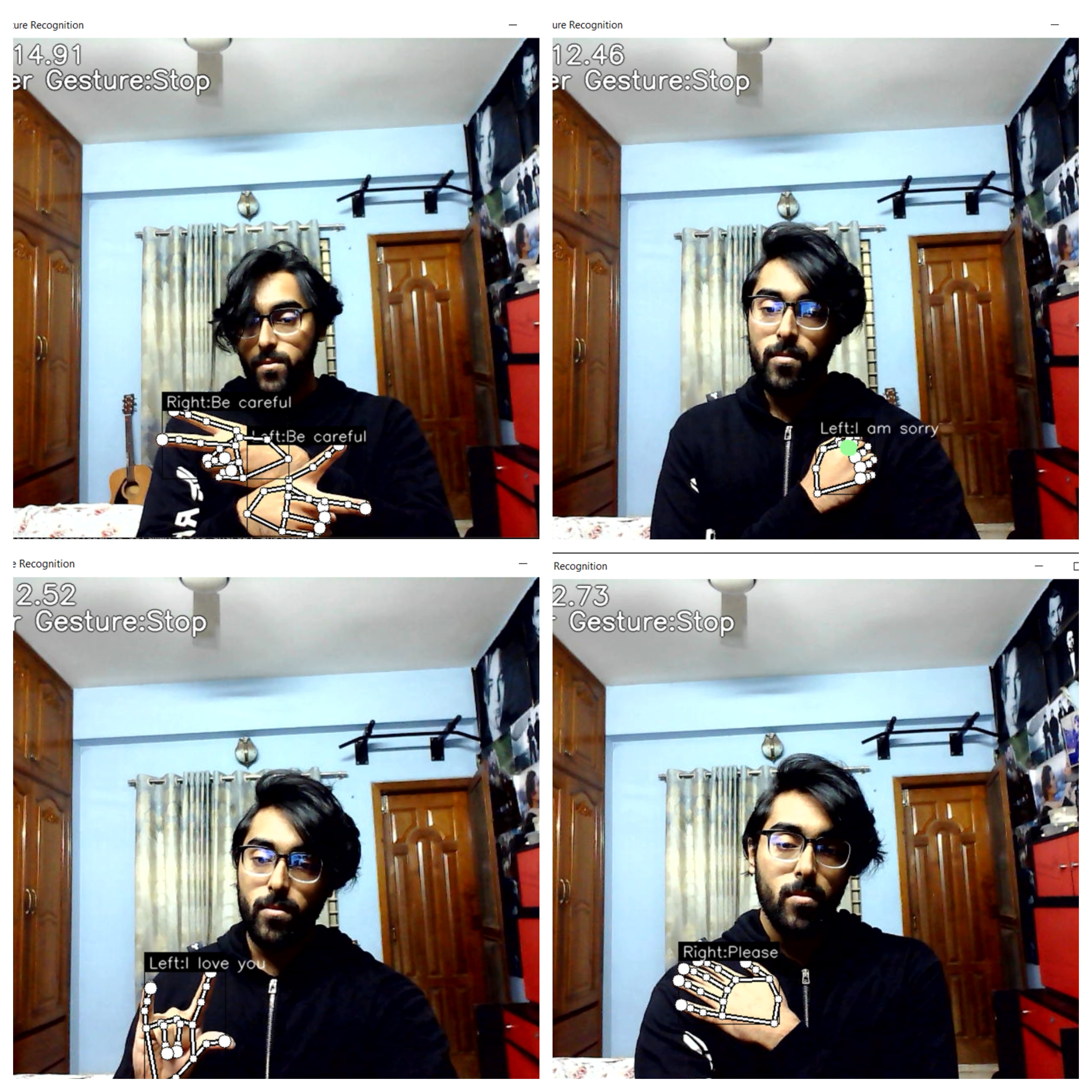}
    \caption{Real time output detection of CNN model
}
    \label{fig:23}
\end{figure}

\noindent
\\
So, from fig. \ref{fig:23}, it can be observed that the model is able to recognize the hand signs performed in real time.

\begin{figure}[!h]
    \centering
    \includegraphics[scale=0.50]{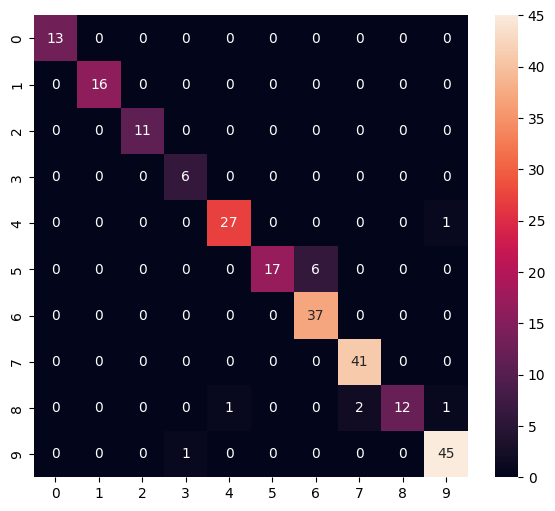}
    \caption{Correlation matrix of CNN model
}
    \label{fig:26}
\end{figure}
\noindent
From the correlation matrix in Fig. \ref{fig:26}, it is clear that all classes are being identified. But sometimes some classes are being falsely identified. For example- ‘Be careful’ (Class-07) has been misclassified as ‘Stop it’ (Class-08) and ‘I need help’ (Class-04) has also been misclassified as ‘Stop it’ (Class-08). So, the model cannot always predict accurately.  

\begin{figure}[!h]
    \centering
    \includegraphics[scale=0.41]{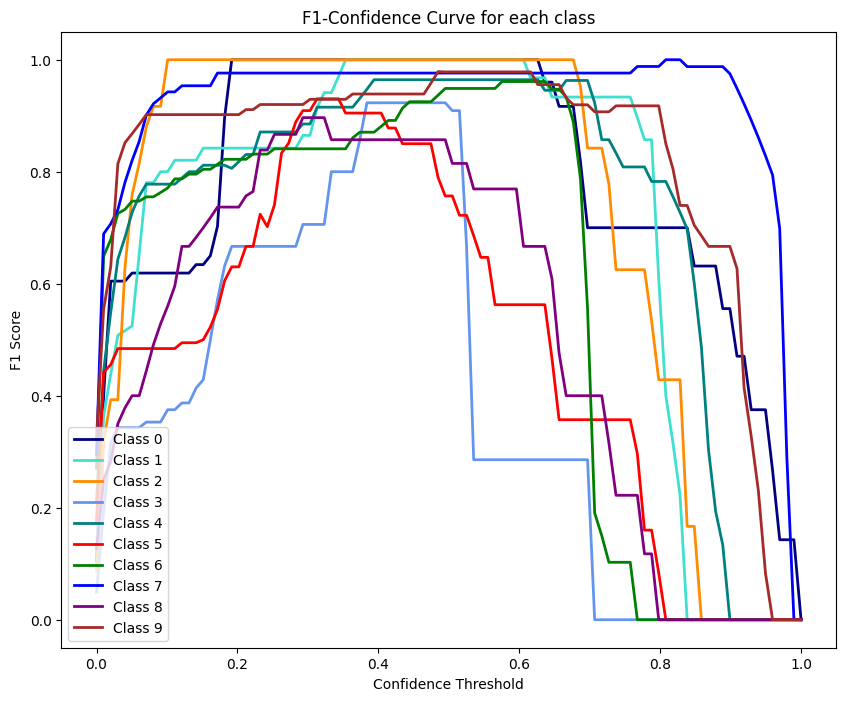}
    \caption{F1 confidence curve of CNN model}
    \label{fig:24}
\end{figure}

\noindent
From the fig. \ref{fig:24}, it can be deduced that most classes have an f1 confidence score on or over 0.8 . One or two classes have a score between 0.7 and 0.8 .\\

\section{Evaluation of results (YOLO models and CNN model)}

From the table in fig. \ref{fig:1278}  it is visible that although the CNN model shows a high accuracy, sometimes it makes false detection. The model is unable to always perfectly identify the hand signs where both hands have been used to perform the sign despite providing enough data. So, it is difficult for this model to identify or recognize comparatively complex hand sign gestures. After seeing its performance, we have come to the conclusion that this model is not suitable for large scale work.
\begin{figure}[!h]
    \centering
    \includegraphics[scale=0.5]{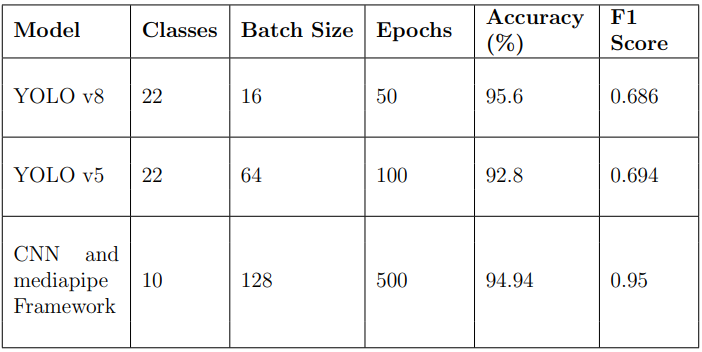}
    \caption{Performance comparison of the models}
    \label{fig:1278}
\end{figure}
\\The YOLO models on the other hand have performed much better than the CNN model. Though the YOLOv5 model shows a little bit less accuracy than the CNN model, it does not do any type of false detection. The YOLOv8 model also does not have any case of making any wrong detection. The YOLOv8 model functions better than the YOLOv5 model. YOLOv5 shows an accuracy of 92.8 percent after training the model for 100 epochs but the YOLOv8 model shows an accuracy of 95.6 percent after training the model for just 50 epochs. Therefore it is evident that the YOLOv8 model is the best one here for real time sign language detection out of all the models used.   

\section{\textbf{Conclusion}}
\noindent
In the text to sign language conversion framework, there are certain sentences that contain stop words (For example - ‘ - apostrophe) that we utilized for filtering are not compatible with the framework.  One further drawback is that the video outputs are not smooth enough. We want to work on these problems in the future and also incorporate a 3D model with smoother transitions.Adding to that,we aim to work with video data on the YOLOv8 model. If we get enough resources to increase the computational power, we will work on a large dataset where there will be hundreds of classes and add also facial expression recognition in our work. In addition to that, we plan to make an app version of this model and framework too. To top it off, we state that our work here on ASL detection can also be applied to other sign languages as well.According to the World Health Organization (WHO), with 1.5 billion people in the world already suffering from hearing loss and the number can increase to over 2.5 billion by 2050.\cite{b56} The deaf community is deprived of basic human rights like health care, education and even minimum wage jobs simply because of their inability to communicate with the hearing people using spoken language. Our YOLO based model and the NLP based framework aim to bridge this communication gap that is prevalent in the community for a long time by providing the fastest real time solution. This will ensure an equal spot for the deaf people in the society by overcoming the language barrier. In conclusion, our system will be helpful for both hearing and hearing impaired people to communicate effectively with one another by shortening the existing communication gap.

\end{document}